\documentclass{article}
\usepackage{iclr2027_conference,times}
\usepackage{xcolor}

\usepackage{amsmath,amsfonts,bm}

\def\eqref#1{equation~\ref{#1}}

\def\1{\bm{1}}

\DeclareMathAlphabet{\mathsfit}{\encodingdefault}{\sfdefault}{m}{sl}
\SetMathAlphabet{\mathsfit}{bold}{\encodingdefault}{\sfdefault}{bx}{n}

\usepackage{hyperref}
\usepackage{url}
\usepackage{graphicx}

\definecolor{cornellred}{rgb}{0.7, 0.11, 0.11}
\definecolor{cadmiumgreen}{rgb}{0.0, 0.42, 0.24}
\definecolor{Blue9}{rgb}{0.098, 0.3, 0.9}

\hypersetup{
    linkcolor = cornellred,
    citecolor = cadmiumgreen,
    colorlinks = true,
    urlcolor = Blue9
}

\title{Future Video Generation Better Aligns with \\ the Human Visual Cortex than\\Observed Video}

\author{
Chang-Bae Bang$^{1,2,3}$ \quad
Hyungjin Chung$^{4\dagger}$ \quad
Byung-Hoon Kim$^{1,2,3,5\dagger}$ \\
\\
$^{1}$Department of Psychiatry, Yonsei University College of Medicine \\
$^{2}$Institute of Behavioral Sciences in Medicine, Yonsei University College of Medicine \\
$^{3}$Department of Biomedical Systems Informatics, Yonsei University College of Medicine \\
$^{4}$Department of Computer Science \& Engineering, Korea University \\
$^{5}$Yonsei Institute for Digital Healthcare, Yonsei University \\
\\
\texttt{changbae.bang@yonsei.ac.kr}, \;
\texttt{hj\_chung@korea.ac.kr}, \;
\texttt{egyptdj@yonsei.ac.kr}
}

\iclrfinalcopy
\begin{document}

\maketitle

{\renewcommand{\thefootnote}{\fnsymbol{footnote}}%
 \footnotetext[2]{Co-corresponding authors.}}

\begin{abstract}
Studying the alignment between the internal representations of vision models and the responses of the visual cortex to the same observed visual stimuli has enabled us to better understand human visual processing. However, studies so far have largely overlooked the fact that the human brain not only processes observed visual stimuli, but also predicts upcoming stimuli based on what has been observed. Accordingly, we hypothesize that internal representations for generating future video frames are better aligned with the predictive nature of human visual processing than representations of the observed video itself. To this end, we compare the alignment between human video-watching fMRI responses in the visual cortex and the internal representations from two types of video diffusion models, an autoregressive (AR) model and its non-AR base model. We first conduct a within-model analysis of the AR video diffusion model and show that the representations for future video generation align better with the visual cortex than the representations of the observed video. We then compare the internal representations of the AR model with those of its non-AR base model and again show that the representations for future video generation align better with the visual cortex than the representations for observed video reconstruction by the base model. Specifically, the alignment of observed video reconstruction is concentrated in lower-order visual cortex, whereas that of future video generation is concentrated in higher-order visual cortex.  Finally, we show in a human behavioral experiment that humans prefer videos generated by amplifying the contributions of individual layers that align better with the visual cortex.
\end{abstract}

\begin{figure}[h]
    \centering
    \includegraphics{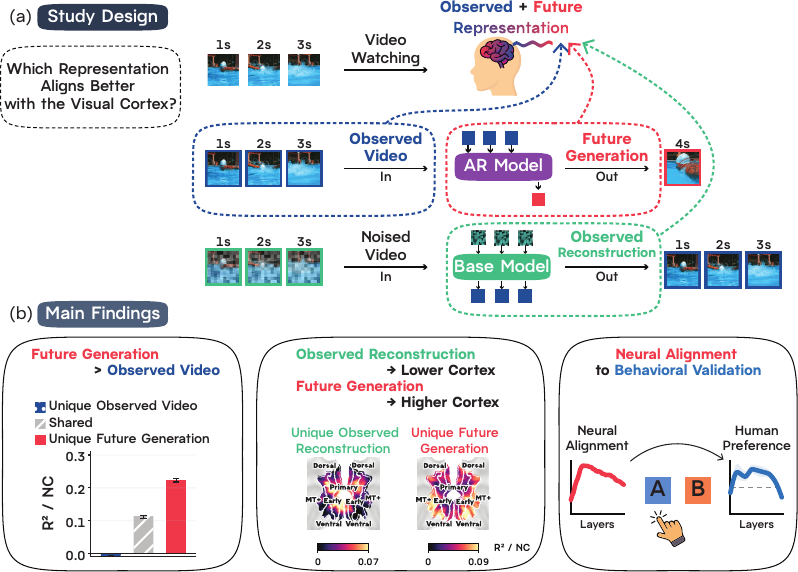}
    \caption{\textbf{Study Design and Main Findings.} (a) We map the representations of the observed video, future video generation, and observed video reconstruction onto the visual cortex. (b) Left, future video generation aligns better with the visual cortex than the observed video. Middle, observed video reconstruction aligns better with lower-order visual cortex, whereas future video generation aligns better with higher-order visual cortex. Right, humans prefer videos generated by amplifying the layers that align better with the visual cortex.}
    \label{fig:overview}
\end{figure}

\section{Introduction}
Studying the representational alignment between vision models and the human brain has enabled us to better understand human visual processing~\citep{khalighrazavi2014, yamins2014, kriegeskorte2018}. 
For example, by examining which features or layers of the model best explain neural responses, we can infer how visual information may be processed in the human brain~\citep{kriegeskorte2008, naselaris2011, schrimpf2018}.
While these brain–vision model representation alignment studies have provided insights into human visual processing, studies to date have largely focused on representations of the observed visual stimulus~\citep{conwell2024, sartzetaki2025, hofling2026}. 
The brain, however, does not only passively respond to the observed stimulus, but also predicts the stimulus that is about to arrive~\citep{raoballard1999, keller2018, hogendoorn2018}. 
Accordingly, we compare the internal representations of a model that generates future video frames with those of the human visual cortex.

Recently, video generative models based on the diffusion transformer~\citep{peebles2022dit} have demonstrated the capability to produce photorealistic videos~\citep{openai2024sora, wan2025}. Conventionally, video diffusion models generate video sequences by progressively denoising latent representations across all frames simultaneously. However, such designs disregard the inherent temporal order of causality. To align with the causal order of time in which visual stimuli naturally unfold, recent approaches have integrated the autoregressive (AR) framework with diffusion models to generate the future video conditioned on the observed video~\citep{yin2024causvid, huang2025selfforcing, yuan2026helios}. The AR model represents how the visual stimulus unfolds over time to generate a future video that is semantically and physically plausible based on the observed video.

We hypothesize that the internal representations for generating future video frames are better aligned with the predictive nature of human visual processing than the representations of the observed video itself. To this end, we compare the internal representations from two types of video diffusion models, an AR model and its non-AR base model. Using human fMRI responses recorded while subjects watch 3\,s naturalistic videos~\citep{lahner2024bmd}, we map the representations of the observed video stimulus, future video generation, and observed video reconstruction (Figure~\ref{fig:overview}a). Moreover, we test whether the alignment of the representation for future video generation is validated at the behavioral level with a human video preference experiment.

\newpage

\paragraph{Contributions (Figure~\ref{fig:overview}b).}
\begin{itemize}
    \item \textbf{We show that future video generation aligns better with the visual cortex than the observed video.}
    The representation for future video generation aligns better than the representation of the observed video
    that both the AR model and the human brain receive.
    \item \textbf{We show that the two representations, for future video generation and observed video reconstruction, align with different
    levels of the visual hierarchy.} The alignment of observed video reconstruction is
    concentrated in lower-order visual cortex, whereas that of future video
    generation is concentrated in higher-order visual cortex.
    \item \textbf{We validate the layer profile of neural alignment in human
    behavior.} Amplifying the contribution of an individual layer during future
    video generation, we show that humans prefer videos generated by amplifying
    the layers that align better with the visual cortex.
\end{itemize}

\section{Related Work}
\subsection{Brain--Vision Model Representation Alignment}
The internal representations of vision models turn out to partly resemble the representations of the human visual cortex~\citep{khalighrazavi2014, yamins2014, schrimpf2018}. Presenting the same visual stimulus to a human and to a vision model, and comparing their internal representations, enables us to understand human visual processing. The internal representations of vision models optimized for object recognition align with the representation in the visual cortex~\citep{yamins2014}, and deeper layers align best with higher-order visual cortex~\citep{guclu2015}. Large-scale recordings of human cortical responses to natural images and videos~\citep{allen2022, lahner2024bmd} have enabled systematic comparisons across a large pool of vision models. These comparisons show that, for images, the visual training diet exerts a larger and more consistent effect on representation alignment than architecture or task objective~\citep{conwell2024}. For videos, temporal modeling drives representation alignment to lower-order visual cortex while the classification task drives alignment to higher-order visual cortex~\citep{sartzetaki2025}. The comparison has also been extended to the internal representations of an image generative model. The U-Net bottleneck of a latent image diffusion model aligns with the visual cortex best at the noisiest denoising step, and as denoising proceeds the first layer aligns better with lower-order visual cortex while the bottleneck aligns better with higher-order visual cortex~\citep{takagi2023}.

\subsection{Predictive Nature of Human Visual Processing}
The human visual cortex represents not only the response to the observed stimulus but also the future stimulus. According to predictive coding theory, the brain uses its prior experience to build an internal generative model of the external world~\citep{raoballard1999, lee2015, keller2018}. With this internal model, the brain can infer the cause of the observed visual stimuli and predict future stimuli. Indeed, the representation of future visual stimuli has been confirmed in human neuroimaging studies with visual tasks. The human brain represents the next position of a moving object before its input arrives~\citep{hogendoorn2018, blom2020}, and when presented with a partial cue, even lower-order visual cortex represents the full sequence of previously experienced visual stimuli~\citep{ekman2017}. Higher-order visual cortex is thought to represent visual stimuli further into the future than lower-order visual cortex while humans observe naturalistic video~\citep{devrieswurm2023, lee2021}.

\subsection{Internal Representation of Diffusion Video Models}
A non-AR diffusion video model, which denoises all frames of the video together, learns an internal representation that carries rich visual information~\citep{xiang2023ddae, li2023zeroshot, yu2025repa, geyer2023tokenflow, nam2025difftrack}. This representation is rich enough to be adapted to vision tasks on the observed video, such as action recognition~\citep{guimaraes2025}, depth estimation~\citep{king2026gen4u}, point tracking~\citep{son2025, nam2025difftrack, yuan2026denoise}, and physical plausibility estimation~\citep{esmati2026}. In contrast to the non-AR model, an AR diffusion video model sequentially generates the future video based on the observed video, allowing video generation to follow the natural temporal order. The AR model represents how the future visual stimulus unfolds over time to generate a future video that is semantically and physically plausible. Indeed, aligning the representation of an AR diffusion video model to the representation of the future video improves its future video generation quality~\citep{yu2026videomirai}.

\section{Preliminaries}
\paragraph{Human Video fMRI Dataset.}
We use the BOLD Moments Dataset \citep{lahner2024bmd}, in which ten subjects watched 1{,}102 3\,s naturalistic videos while whole-brain fMRI was recorded, 1{,}000 of them for training and 102 for testing. Preprocessing follows the MOSAIC pipeline \citep{lahner2025mosaic} and leaves one response per video at each cortical vertex (Appendix~\ref{app:dataset}),
\begin{equation}
Y \in \mathbb{R}^{N \times P}, \quad
m \in \{1,\dots,10\}, \quad
p \in \{1,\dots,P\}, \quad
r \subseteq \{1,\dots,P\},
\label{eq:notation}
\end{equation}
where $Y$ is the response of one subject over $N$ videos and $P$ vertices, $m$ indexes subjects, $p$ indexes vertices, $r$ is a region composed of a group of vertices, and $|r|$ is the number of vertices it contains.

\paragraph{Diffusion Video Generative Model.}
A diffusion video generative model generates a video by iteratively denoising noisy video tokens. A variational autoencoder $\mathcal{E}$ maps a video $v$ to a latent $z_0 = \mathcal{E}(v) \in \mathbb{R}^{C \times T \times H \times W}$, and generation follows a flow-matching path between noise and data~\citep{lipman2023}. A latent on this path at noise level $\sigma$ interpolates the clean latent $z_0$ with Gaussian noise,
\begin{equation}
z_\sigma = (1-\sigma)\,z_0 + \sigma\,\epsilon, \qquad \epsilon \sim \mathcal{N}(0, I).
\label{eq:interp}
\end{equation}
We order the schedule $\sigma_0 > \sigma_1 > \dots > \sigma_{S-1}$, so that step $s = 0$ is the noisiest and step $s = S-1$ the cleanest. The transformer has $B$ layers, numbered $b = 0,\dots,B-1$. Let $H_{s,b}$ denote the hidden states that layer $b$ produces at noise level $\sigma_s$, and let $H^{\mathrm{in}}_{s}$ denote the sequence that enters the first layer.

\section{Methods and Experiment Setup}
\paragraph{Human Brain Data.}
Every vertex is labeled with the HCP-MMP atlas \citep{glasser2016}. Our primary focus is the Visual System, 7{,}894 vertices grouped into five divisions. The five divisions are Primary Visual, Early Visual, Dorsal Stream, Ventral Stream, and MT+ Complex, and a region $r$ is either one division or the Visual System as a whole. Only vertices with a noise ceiling $\mathrm{NC} > 10\%$ enter the analysis. 

\paragraph{Models.}
The main analysis uses helios-base-v2v \citep{yuan2026helios} as the AR model, a 40-layer autoregressive video model fine-tuned from the non-AR base model Wan-2.1 T2V-14B \citep{wan2025}. We replicate the analysis on two further AR models, Self-Forcing \citep{huang2025selfforcing} and CausVid \citep{yin2024causvid}, both 30-layer autoregressive video models fine-tuned from the non-AR base model Wan-2.1 T2V-1.3B.

\paragraph{Video Preparation.}
Every BOLD Moments Dataset video is 3\,s long and square. For helios-base-v2v and Wan-2.1 T2V-14B we resample each clip to $272 \times 272$ at 33\,fps. For Self-Forcing and CausVid we fit it inside an $832 \times 480$ frame, keeping the aspect ratio and padding the sides with black, and sample it at 15\,fps.

\paragraph{Autoregressive Model.}
We use an AR model built on a diffusion video generative model that generates future video frames based on observed video stimuli. The AR model generates the future video by denoising future video tokens. Let $z^{\mathrm{obs}}$ denote the clean latent of the observed video and $z^{\mathrm{fut}}_{\sigma_s}$ the latent of the future video at noise level $\sigma_s$. The input sequence for every step carries both observed and future tokens at once,
\begin{equation}
H^{\mathrm{in}}_{s} = \big[\;
\underbrace{\mathrm{patch}(z^{\mathrm{obs}})}_{\text{observed video tokens, clean}}
\;;\;
\underbrace{\mathrm{patch}(z^{\mathrm{fut}}_{\sigma_s})}_{\text{future video tokens at noise level } \sigma_s}
\;\big],
\label{eq:arseq}
\end{equation}
with the observed tokens fed in clean at every step and the future tokens carried over from the previous step as they are denoised. Let $c^{\mathrm{obs}} = \mathrm{patch}(z^{\mathrm{obs}})$ denote the observed video embedding, which is the same at every step. The layer activation of the AR model at step $s$ and layer $b$ splits into an observed part and a future part,
\begin{equation}
H^{\mathrm{ar}}_{s,b} = \big[\, H^{\mathrm{obs}}_{s,b} \;;\; H^{\mathrm{fut}}_{s,b} \,\big].
\label{eq:arsplit}
\end{equation}
We extract future tokens $H^{\mathrm{fut}}_{s,b}$, so that each video yields a grid of representations indexed by the denoising step $s = 0,\dots,S-1$ and the layer $b = 0,\dots,B-1$.

\paragraph{Future Video Generation with AR Model.}
For each AR model we feed the 3\,s observed video and let the model generate the 1\,s future video that follows. Because subjects only observed the 3\,s video with no instruction, we leave the text prompt empty. The AR model helios-base-v2v runs 50 denoising steps, and Self-Forcing and CausVid run 4 steps, the default for each model. For each model we extract $c^{\mathrm{obs}}$ once, and $H^{\mathrm{fut}}_{s,b}$ at every step and layer.

\paragraph{Base Model.}
We use a non-AR base model built on a diffusion generative model that reconstructs noised video stimuli. The base model reconstructs the observed video by denoising observed video tokens. We noise $z^{\mathrm{obs}}$ by Equation~\ref{eq:interp} and run one forward pass at each of the noise levels $\sigma_s$, so that at every noise level the base model reconstructs the observed video from its noised form. Let $H^{\mathrm{rec}}_{s,b}$ denote the layer activation of the base model at step $s$ and layer $b$. We extract $H^{\mathrm{rec}}_{s,b}$, which gives a grid over noise levels and layers of the same shape as $H^{\mathrm{fut}}_{s,b}$.

\paragraph{Observed Video Reconstruction with Base Model.}
For each base model we feed the 3\,s observed video noised at the same schedule as its AR model. At each noise level we run one forward pass to denoise the observed video with no text prompt. For each model we extract $H^{\mathrm{rec}}_{s,b}$ at every step and layer.

\paragraph{Vertex-Wise Encoding Model.}
We map three representations onto the cortical response recorded while subjects observed the videos (Figure~\ref{fig:overview}a). The first is the observed video embedding $c^{\mathrm{obs}}$ that the AR model receives (Equation~\ref{eq:arseq}). The second is the representation for future video generation $H^{\mathrm{fut}}_{s,b}$ of the AR model (Equation~\ref{eq:arsplit}). The third is the representation for observed video reconstruction $H^{\mathrm{rec}}_{s,b}$ of the base model. Each of them is flattened and reduced to $d = 6004$ dimensions with a sparse random projection (Appendix~\ref{app:pipeline}), which gives the feature matrices $X^{\mathrm{obs}}$, $X^{\mathrm{fut}}_{s,b}$, and $X^{\mathrm{rec}}_{s,b}$, each in $\mathbb{R}^{N \times d}$ with one row per video. For each representation we fit a vertex-wise encoding model that predicts the response of one cortical vertex~\citep{naselaris2011}. Let $X$ denote any one of these feature matrices. For vertex $p$ we fit ridge regression weights,
\begin{equation}
w_p = \arg\min_{w \in \mathbb{R}^{d}} \big\lVert Y_{:,p} - X w \big\rVert_2^2 + \lambda_p \lVert w \rVert_2^2 ,
\label{eq:ridge}
\end{equation}
with a ridge penalty $\lambda_p$ chosen for that vertex, and take $\hat{Y}_{:,p} = X w_p$ as the predicted response of that vertex. The penalty is chosen by nested cross-validation on the training videos, following the stacked-regression scheme of \citet{lin2024stacking}, and for the two grid representations, $H^{\mathrm{fut}}_{s,b}$ and $H^{\mathrm{rec}}_{s,b}$, the best step and layer are selected on the training folds (Appendix~\ref{app:pipeline}). The test videos take no part in fitting or in this selection, and every result we report is measured on test videos.

\paragraph{Encoding Accuracy.}
Encoding accuracy at a vertex is the coefficient of determination $R^2_p$ of the test-set predictions. Every vertex is assigned a noise ceiling $\mathrm{NC}_p$, the upper bound on how well any model can predict its response given the noise in the fMRI response (Appendix~\ref{app:dataset}). We report the noise-normalized encoding accuracy
\begin{equation}
\tilde{R}^2_p = R^2_p / \mathrm{NC}_p .
\label{eq:acc}
\end{equation}
The encoding accuracy of a region $r$ is its mean over the vertices of the region, $\tilde{R}^2_r = \frac{1}{|r|}\sum_{p \in r} \tilde{R}^2_p$.

\paragraph{Variance Partitioning.}
Two representations can explain the same part of the cortical response. To separate what each explains on its own from what the two explain in common, we fit a joint encoding model that combines representations $a$ and $b$ and measure the unique contribution of each representation. Let $\tilde{R}^2_{\mathrm{stack}}$ be the encoding accuracy of the joint model, and $\tilde{R}^2_a$ and $\tilde{R}^2_b$ the encoding accuracies of $a$ and $b$ alone~\citep{lin2024stacking},
\begin{equation}
U_a = \tilde{R}^2_{\mathrm{stack}} - \tilde{R}^2_b, \qquad
U_b = \tilde{R}^2_{\mathrm{stack}} - \tilde{R}^2_a, \qquad
U_{ab} = \tilde{R}^2_a + \tilde{R}^2_b - \tilde{R}^2_{\mathrm{stack}} .
\label{eq:vpmain}
\end{equation}
$U_a$ and $U_b$ are the unique contributions of the two representations and $U_{ab}$ is their shared contribution. The joint model is explained in Appendix~\ref{app:vp}. We partition two pairs. The within-model pair sets the observed video against future video generation, $(X^{\mathrm{obs}}, X^{\mathrm{fut}})$. The cross-model pair sets observed video reconstruction against future video generation, $(X^{\mathrm{rec}}, X^{\mathrm{fut}})$. Here $X^{\mathrm{fut}}$ and $X^{\mathrm{rec}}$ are the grid cells $X^{\mathrm{fut}}_{s,b}$ and $X^{\mathrm{rec}}_{s,b}$ at the best-aligned step and layer, selected on the training folds for each region and subject (Appendix~\ref{app:pipeline}, Equation~\ref{eq:cellsel}).

\section{Results}

\begin{figure}[t]
    \centering
    \includegraphics{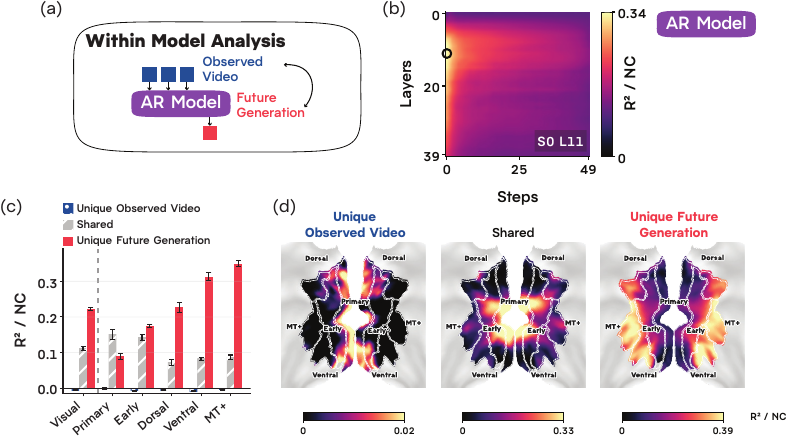}
    \caption{\textbf{Within-model analysis.} (a) The AR model receives the 3\,s observed video and generates the 1\,s future video. (b) Encoding accuracy of the representation for future video generation of the AR model over denoising steps and layers in the Visual System. (c) Unique and shared contributions of the representation for future video generation against the observed video in the Visual System and its five divisions. (d) The same contributions plotted on the cortical surface.}
    \label{fig:within}
\end{figure}

\subsection{Within-Model Analysis}
\label{sec:within}

The AR model receives the 3\,s observed video and generates the 1\,s future video. We map both the representation of the observed video $X^{\mathrm{obs}}$ and the representation for future video generation $X^{\mathrm{fut}}_{s,b}$ onto the visual cortex, and then use variance partitioning to measure the unique contribution of each representation. We report Helios here and repeat the same analysis on CausVid and Self-Forcing in Appendix~\ref{app:replication}.

\paragraph{Future video generation aligns better with the visual cortex than the observed video.}
The representation for future video generation aligns best with the Visual System at step~0 and layer~11, the cell selected on the training folds (Figure~\ref{fig:within}b). In the Visual System the unique contribution of future video generation is $0.222$, the shared contribution is $0.111$, and the unique contribution of the observed video is $-0.006$ (Figure~\ref{fig:within}c). The unique contribution of future video generation is above the unique contribution of the observed video in every division ($p = 0.002$, Appendix~\ref{sec:stats}). Only in Primary Visual does the shared contribution, $0.151$, exceed the unique contribution of future video generation, $0.089$. In every other division the unique contribution of future video generation exceeds both the unique contribution of the observed video and the shared contribution. The shared contribution and the unique contribution of future video generation are $0.142$ and $0.175$ in Early Visual, $0.072$ and $0.227$ in Dorsal Stream, $0.083$ and $0.314$ in Ventral Stream, and $0.087$ and $0.350$ in MT+ Complex.  On the cortical surface the unique contribution of future video generation grows continuously toward higher-order visual cortex, whereas the shared contribution and the unique contribution of the observed video are concentrated in lower-order visual cortex (Figure~\ref{fig:within}d).

\begin{figure}[t]
    \centering
    \includegraphics{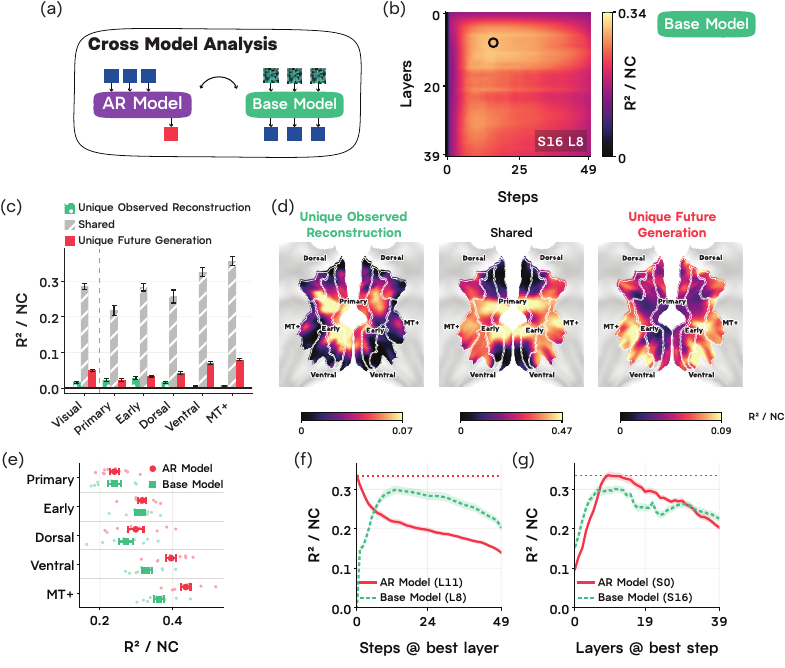}
    \caption{\textbf{Cross-model analysis.} (a) The AR model generates the 1\,s future video and its base model reconstructs the 3\,s noised observed video. (b) Encoding accuracy of the representation for observed video reconstruction of the base model over denoising steps and layers in the Visual System. (c) Unique and shared contributions of future video generation against observed video reconstruction in the Visual System and its five divisions. (d) The same contributions plotted on the cortical surface. (e) Encoding accuracy of the representations for future video generation and observed video reconstruction in the five divisions. (f, g) Encoding accuracy of the two representations along the step axis at each representation's own selected layer and along the layer axis at each representation's own selected step.}
    \label{fig:cross}
\end{figure}

\subsection{Cross-Model Analysis}
\label{sec:cross}

The base model reconstructs the 3\,s observed video from its noised form. We map the representation for observed video reconstruction $X^{\mathrm{rec}}_{s,b}$ onto the visual cortex, and then partition the variance with the representation for future video generation $X^{\mathrm{fut}}_{s,b}$ (Figure~\ref{fig:cross}a). We report Helios here and repeat the same analysis on CausVid and Self-Forcing in Appendix~\ref{app:replication}.

\paragraph{Future video generation aligns better with the visual cortex than observed video reconstruction.}
The representation for observed video reconstruction peaks at step~16 and layer~8, the cell selected on the training folds (Figure~\ref{fig:cross}b). Its best encoding accuracy, $0.301$, is below the best encoding accuracy of future video generation, $0.334$. In Primary Visual and Early Visual the encoding accuracies of future video generation and observed video reconstruction are similar, $0.240$ and $0.240$ in Primary Visual and $0.317$ and $0.310$ in Early Visual (Figure~\ref{fig:cross}e). However, future video generation aligns better in higher-order visual cortex. In Dorsal Stream future video generation is at $0.299$ and observed video reconstruction at $0.272$, in Ventral Stream $0.396$ and $0.329$, and in MT+ Complex $0.436$ and $0.363$ ($p = 0.002$ in all three).

\paragraph{Observed video reconstruction aligns with lower-order visual cortex, and future video generation aligns with higher-order visual cortex.}
In Primary Visual the two unique contributions are $0.023$ for observed video reconstruction and $0.022$ for future video generation, in Early Visual $0.027$ and $0.034$, in Dorsal Stream $0.015$ and $0.042$, in Ventral Stream $0.004$ and $0.071$, and in MT+ Complex $0.006$ and $0.080$ (Figure~\ref{fig:cross}c). On the cortical surface the unique contribution of future video generation grows continuously toward higher-order visual cortex, whereas the unique contribution of observed video reconstruction is concentrated in lower-order visual cortex (Figure~\ref{fig:cross}d).

\paragraph{Alignment of observed video reconstruction peaks at a less noisy step than future video generation.}
The alignment profiles of observed video reconstruction and future video generation differ along the step axis (Figure~\ref{fig:cross}f). The alignment of future video generation is highest at the noisiest step and decreases monotonically as denoising proceeds, whereas the alignment of observed video reconstruction peaks early in the noise schedule and declines after it. Along the layer axis the alignment of both representations follows an inverted U with the maximum at an early-to-middle layer (Figure~\ref{fig:cross}g).

\begin{figure}[t]
    \centering
    \includegraphics{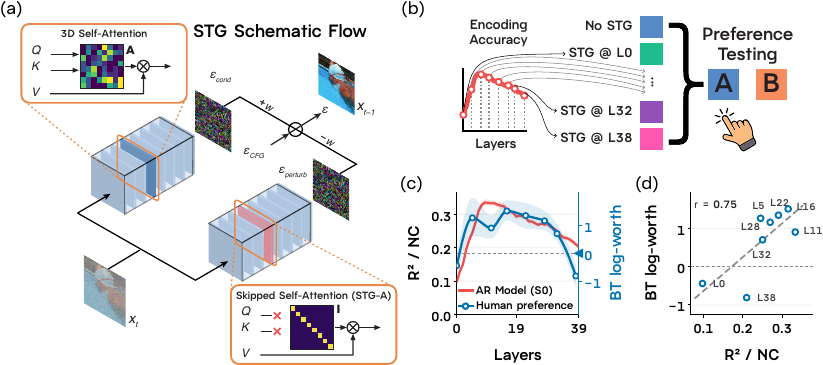}
    \caption{\textbf{Layer-wise guidance and human preference.} (a) STG skips the
    self-attention of one layer and pushes the sample away from the perturbed prediction
    (Equation~\ref{eq:stg}). The schematic follows the layout of \citet{ahn2024pag}.
    (b) Pairwise preference test among the nine arms, the unguided arm (No STG) and eight STG-guided arms. (c) Encoding accuracy of the AR model at step~0 in the Visual System and human preference as Bradley--Terry log-worth along the layer axis. (d) Encoding accuracy against human preference, with the Pearson $r$.}
    \label{fig:behav}
\end{figure}

\subsection{Neural Alignment to Behavioral Validation}
\label{sec:pref}
We finally ask whether the neural alignment of future video generation extends to human behavior. Spatiotemporal skip guidance (STG; \citealp{hyung2025stg}) amplifies the contribution of a layer during video generation (Figure~\ref{fig:behav}a, Appendix~\ref{app:stg}). We test whether the neural alignment of future video generation is validated by human preference for videos generated under layer-wise amplification with STG. We apply STG to one layer $b \in \{0, 5, 11, 16, 22, 28, 32, 38\}$ at a time while the AR model generates a 5\,s future video conditioned on the observed 3\,s video. We then run A/B tests on the combined 8\,s videos with five human observers, 720 preference trials in total (Figure~\ref{fig:behav}b, Appendix~\ref{app:humaneval}).

\paragraph{Humans prefer videos generated by amplifying well-aligned layers.}
The neural alignment of future video generation and human preference along the layer axis both follow an inverted U (Figure~\ref{fig:behav}c). Among the eight guided layers, encoding accuracy at step~0 is highest at layer~11 and preference at layer~16. Over the eight layers the two measures correlate at $r = 0.75$ ($p = 0.029$, Figure~\ref{fig:behav}d).

\section{Discussion}
\paragraph{Future video generation aligns better than the observed video.}
The representation for future video generation aligns better with the visual cortex than the observed video itself. Moreover, the observed tokens themselves add no unique contribution beyond future video generation in any division (Appendix~\ref{app:ablation}), which further confirms that future video generation is the key to brain alignment. This may be indirect evidence that the visual cortex performs future-oriented computation, especially in higher-order visual cortex, where human experiments have also shown predictive signals~\citep{lee2021, devrieswurm2023, tarderstoll2024}.

\paragraph{Why the noisiest step and an early-to-middle layer.}
The representation for future video generation aligns best at the noisiest step and an early-to-middle layer. As denoising goes on, the future video settles into one of the many futures that could follow. Interestingly, this uncommitted state aligned best with the visual cortex. This finding is in line with Bayesian theories of neural coding, where the activity of a neural population encodes the whole range of possible stimuli rather than a single-valued prediction~\citep{ma2006ppc, vanbergen2019}. On the other hand, the representation for observed video reconstruction aligns best at a less noisy step. Moreover, the unique contribution of observed video reconstruction is concentrated in lower-order visual cortex, which processes simple visual features such as edges, contours, and textures~\citep{rolls2024}. As such fine details are refined in the later stages of the denoising schedule~\citep{wang2023painters}, these results suggest that, compared with future video generation, observed video reconstruction focuses more on refining such fine details than on grasping the global semantics of the video. Along the layer axis, prior work reports that the visual representation is rich in the intermediate layers of diffusion transformers~\citep{nam2025difftrack, yuan2026denoise, yu2025repa} and poor in the first and last layers~\citep{yuan2026denoise}, which agrees with our finding.

\paragraph{Comparison with a wide range of vision models.}
Previous studies compared the internal representations of vision models of various architectures, training datasets, and objectives to understand human visual processing~\citep{conwell2024, sartzetaki2025}. We also compared the internal representations of a wide range of vision models with the representation for future video generation (Appendix~\ref{app:encoders}). The internal representations of self-supervised video models aligned best with the visual cortex, as reported before~\citep{hofling2026, tang2025}. However, the gap in brain alignment between the representation for future video generation and the representations of self-supervised video models narrowed toward higher-order visual cortex. This may be further indirect evidence that higher-order visual cortex implements a mechanism that predicts future visual stimuli.

\paragraph{From neural alignment to human preference.}
STG guides the video generation path away from a perturbed weak model in which the self-attention of one layer is skipped~\citep{hyung2025stg}. Applying STG to a layer amplifies the representation of that layer. In our experiment, humans preferred the future videos generated by amplifying the layers that aligned better with the visual cortex. A layer that aligns better with the visual cortex may hold a richer visual representation, and amplifying it may make the generated video more natural to a human. This is in line with previous studies showing that, for diffusion-based generative models, aligning the internal representation to a richer visual representation improves the quality of the generated visual contents~\citep{yu2025repa, zhang2025videorepa}.

\paragraph{Limitations.}
First, our analysis framework, which feeds a 3\,s observed video and generates the 1\,s future video that follows, limits understanding of how the brain represents the past and the future over a longer context and a longer horizon~\citep{hasson2008, lee2021, brunec2022, tarderstoll2024}. Second, the cortical response evolves over time while the subject watches the video~\citep{devrieswurm2023}, but we map one response per video onto the visual cortex and do not compare the time course of the response with the internal representation of the AR model. Finally, our behavioral experiment shows that the layer profile of neural alignment is reflected in human preference, but it does not provide a causal account of how neural alignment gives rise to preference.

\section{Conclusion}
In this study, we examined the representational alignment of an AR and a non-AR video generative model with the visual cortex. We tested which of the three representations, the observed video, future video generation, and observed video reconstruction, aligns best with the visual cortex. With vertex-wise encoding and variance partitioning, we showed that future video generation aligns better than both the observed video and its reconstruction. More specifically, the alignment advantage of the representation for future video generation grows along the visual hierarchy. In contrast, the unique contributions of the observed video and its reconstruction are concentrated in lower-order visual cortex. Finally, by amplifying one layer during future video generation, we showed that humans preferred the videos generated from the layers that align well with the visual cortex.

\subsection*{AI use statement}

In this work, we used generative AI tools to provide feedback on the research methodology and experiments, to implement methods, to assist with translation, and to interpret results. We have not used generative AI tools to generate synthetic datasets, to develop theoretical models or conceptual frameworks, to formulate mathematical claims, to provide ingredients for or write mathematical proofs, to propose or refine hypotheses, to clean or reformat datasets, or to support qualitative or thematic data analysis. Additionally, we used generative AI tools to create and edit software code, to draft parts of the paper, to summarize existing literature, to identify relevant literature and search for information, for brainstorming, to edit the paper to improve readability, and to format references. We have reviewed all AI-assisted work. LLM-generated code was verified and tested by the authors, and all AI-assisted text, translations, and references were checked and revised by the authors. We take responsibility for the final content of this work, including text, claims, or artifacts produced with the aid of generative AI.

\subsection*{Ethics statement}
The fMRI data are from the publicly released BOLD Moments Dataset and are used in their anonymized form. The human preference experiment was a video-watching task with no risk to the observers, who were adult volunteers and gave informed consent, and no identifying information about them was collected or reported.

\bibliography{refs}
\bibliographystyle{iclr2027_conference}

\appendix
\section{Appendix}
\subsection{Replication on Other Autoregressive Video Models}
\label{app:replication}
\begin{figure}[h]
    \centering
    \includegraphics{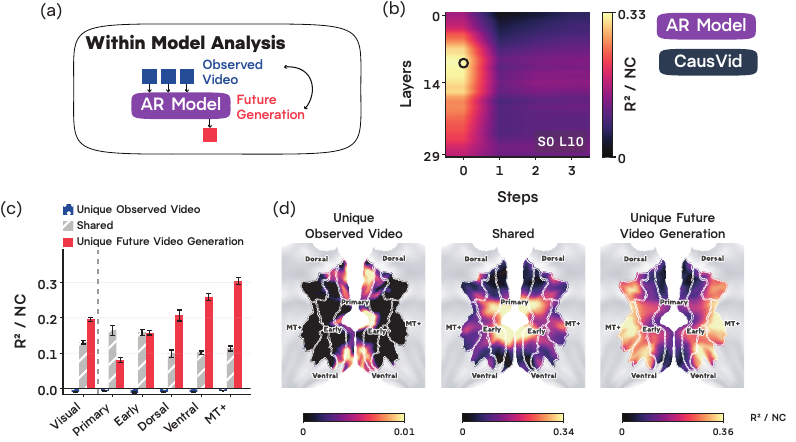}
    \caption{\textbf{Within-model analysis of CausVid.} }
    \label{fig:repl_within_causvid}
\end{figure}

\begin{figure}[h]
    \centering
    \includegraphics{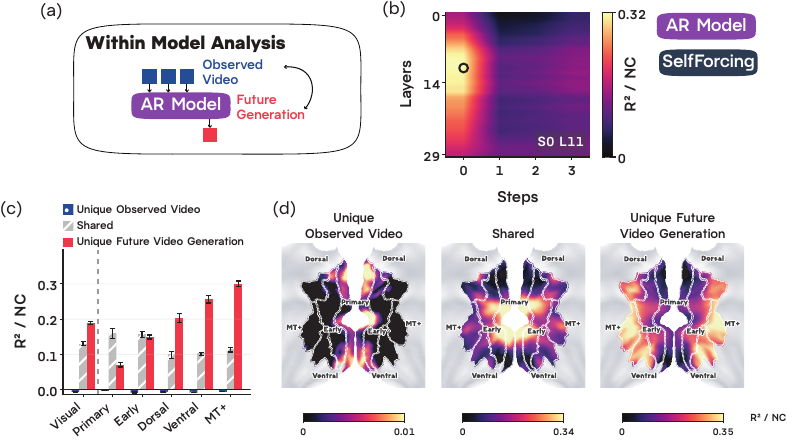}
    \caption{\textbf{Within-model analysis of Self-Forcing.}}
    \label{fig:repl_within_selfforcing}
\end{figure}

\begin{figure}[h]
    \centering
    \includegraphics[width=0.8\linewidth]{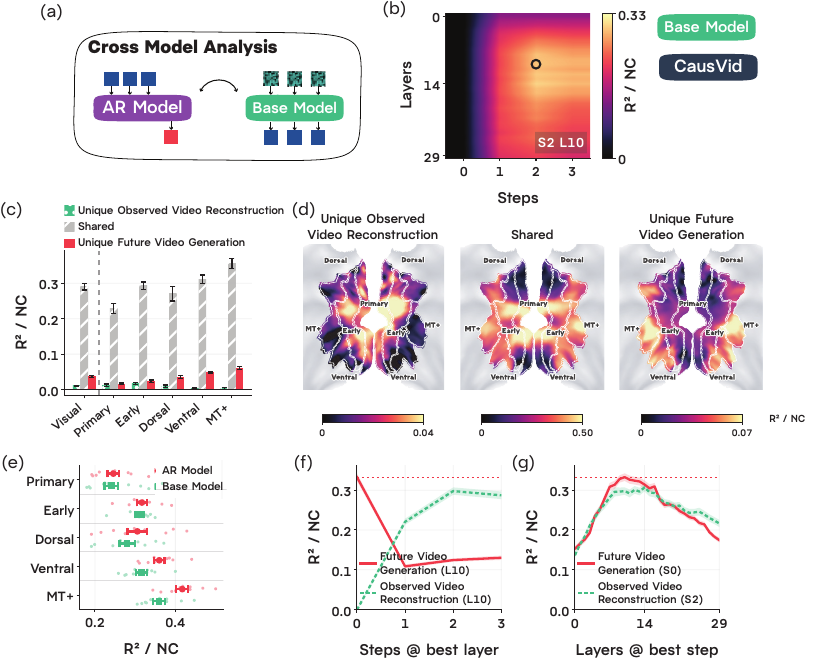}
    \caption{\textbf{Cross-model analysis of CausVid.}}
    \label{fig:repl_cross_causvid}
\end{figure}

\begin{figure}[h]
    \centering
    \includegraphics[width=0.8\linewidth]{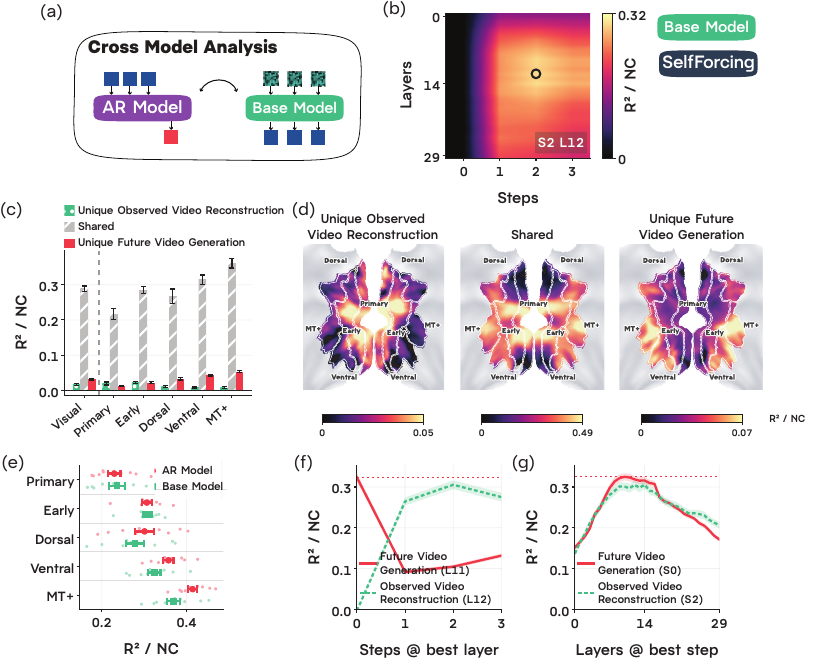}
    \caption{\textbf{Cross-model analysis of Self-Forcing.}}
    \label{fig:repl_cross_selfforcing}
\end{figure}

We repeat the within-model and cross-model analyses on two other AR video models, CausVid~\citep{yin2024causvid} and Self-Forcing~\citep{huang2025selfforcing}. Both are fine-tuned from the non-AR base model Wan-2.1 T2V-1.3B and generate the future video in four denoising steps. We extract the representations for future video generation and for observed video reconstruction from all four steps and all 30 layers, and run the same encoding and variance partitioning as in the main analysis (Figures~\ref{fig:repl_within_causvid}--\ref{fig:repl_cross_selfforcing}).

\subsection{Ablation Study}
\label{app:ablation}
\paragraph{Token Readout.}
The activation of the AR model while it generates the future video has an observed part and a future part, $H^{\mathrm{ar}}_{s,b} = [\, H^{\mathrm{obs}}_{s,b} \;;\; H^{\mathrm{fut}}_{s,b} \,]$ (Equation~\ref{eq:arsplit}). In the main analysis we read out the future part $H^{\mathrm{fut}}_{s,b}$. For the ablation we also read out the observed part $H^{\mathrm{obs}}_{s,b}$ and the concatenation $[\, H^{\mathrm{obs}}_{s,b} \;;\; H^{\mathrm{fut}}_{s,b} \,]$, and apply the same sparse random projection and vertex-wise ridge regression as in the main analysis.

\begin{figure}[h]
    \centering
    \includegraphics{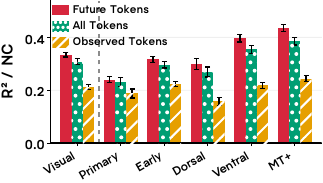}
    \caption{\textbf{Encoding accuracy of future video generation under different token readouts.}}
    \label{fig:ablation}
\end{figure}
\paragraph{Future Tokens Align Best with the Visual Cortex.}
Reading out only the future tokens aligns best with the visual cortex (Figure~\ref{fig:ablation}). In the Visual System the encoding accuracy of the future tokens reaches $\tilde{R}^2 = 0.334$ against $0.211$ for the observed tokens, with the concatenated tokens between them at $0.309$.

\paragraph{Variance Partitioning against the Observed Tokens.}
The observed tokens $H^{\mathrm{obs}}_{s,b}$ also form a grid over denoising steps and layers, so we partition the variance between $H^{\mathrm{obs}}_{s,b}$ and $H^{\mathrm{fut}}_{s,b}$. The cell of each representation is selected on the training folds for each region and subject. 
\begin{figure}[h]
    \centering
    \includegraphics{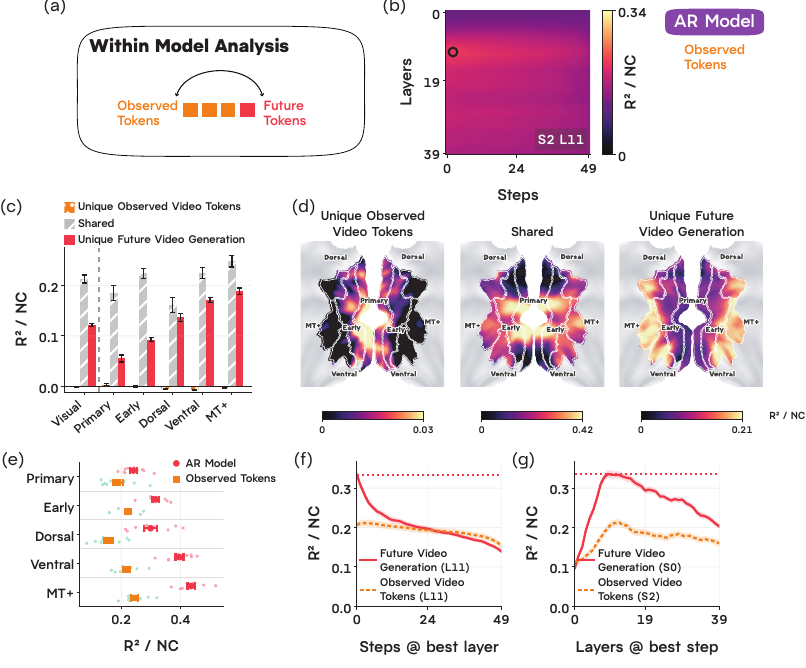}
    \caption{\textbf{Within-model analysis against the observed tokens.}}
    \label{fig:ablation_hobs}
\end{figure}
\paragraph{Future Tokens Keep Their Unique Contribution against the Observed Tokens.}
The observed tokens align best with the Visual System at step~2 and layer~11 (Figure~\ref{fig:ablation_hobs}b). Again, future video generation aligns better than the observed tokens. In the Visual System the shared contribution is $0.213$, the unique contribution of future video generation is $0.121$, and the unique contribution of the observed tokens is $-0.001$ (Figure~\ref{fig:ablation_hobs}c). The unique contribution of future video generation is above the unique contribution of the observed tokens in every division ($p = 0.002$), and it grows along the hierarchy, from $0.055$ in Primary Visual and $0.093$ in Early Visual to $0.137$ in Dorsal Stream, $0.171$ in Ventral Stream, and $0.189$ in MT+ Complex. On the cortical surface the unique contribution of future video generation is again concentrated in higher-order visual cortex (Figure~\ref{fig:ablation_hobs}d).

\subsection{Comparison with a Wide Range of Vision Models}
\label{app:encoders}
We compared the internal representations of a wide range of vision models with the representation for future video generation of the AR model. We included both video and image vision models, 80 models in total.

\begin{figure}[h]
    \centering
    \includegraphics{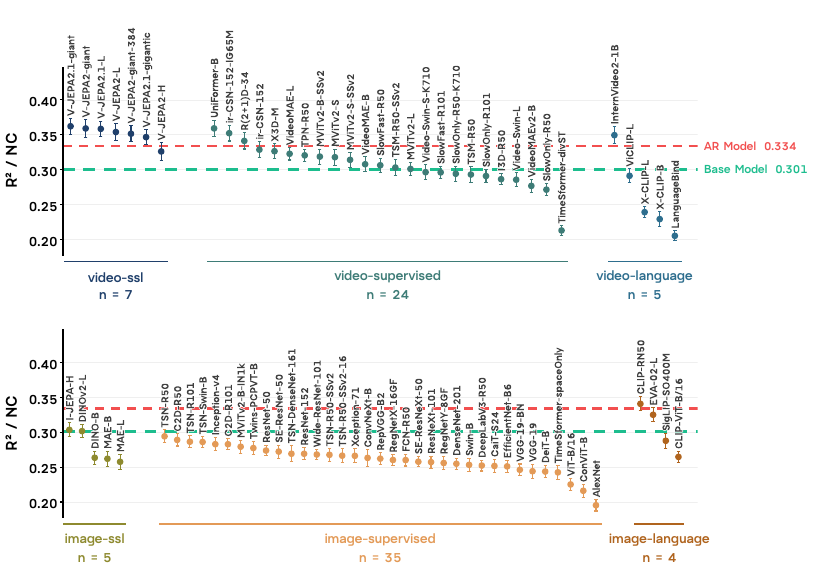}
    \caption{\textbf{Encoding accuracy of vision models in the Visual System.} }
    \label{fig:encoder_strip}
\end{figure}

\begin{figure}[h]
    \centering
    \includegraphics{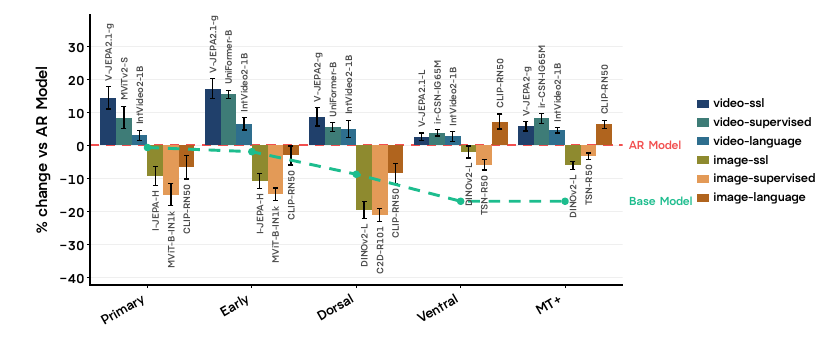}
    \caption{\textbf{Best vision model of each group in each division.} }
    \label{fig:encoder_summary}
\end{figure}

\paragraph{Vision Model Activation Extraction.}
A video vision model that takes $n$ frames spaced $\delta$ frames apart sees $n\delta$ of the $T$ frames of a video in one pass, so we run $K = \lceil T / (n\delta) \rceil$ passes $v^{(1)},\dots,v^{(K)}$ to cover the whole video. An image vision model has no temporal window, so a pass is a single frame and $K = T$. Let $h_\ell$ denote the activation at layer $\ell$ of a vision model, where $\ell = 1,\dots,L$ indexes the composite stages of the backbone. The representation of the vision model $X^{\mathrm{vis}}_{\ell}$ holds one row per video, the layer activation flattened over tokens and channels and averaged over passes,
\begin{equation}
\big[X^{\mathrm{vis}}_{\ell}\big]_{v} \;=\; \frac{1}{K}\sum_{k=1}^{K} \operatorname{vec} h_\ell\big(v^{(k)}\big).
\label{eq:encread}
\end{equation}
Vision models have no denoising axis, so their grid reduces to the layer axis alone. $X^{\mathrm{vis}}_{\ell}$ goes through the same sparse random projection and the same nested cross-validation (Appendix~\ref{app:pipeline}). Figure~\ref{fig:encoder_strip} shows the encoding accuracy of every model in the Visual System, and Figure~\ref{fig:encoder_summary} the best model of each group in each division.

\subsection{fMRI Dataset and Preprocessing}
\label{app:dataset}

\paragraph{Stimuli and Acquisition.}
The BOLD Moments Dataset records whole-brain fMRI responses of ten subjects to 1{,}102 naturalistic videos of 3\,s~\citep{lahner2024bmd}. Each video was shown at a visual angle of 5$^\circ$ while the subject held central fixation, and each trial was the 3\,s video followed by a 1\,s inter-trial interval. The 1{,}000 training videos were presented three times each and the 102 test videos ten times each. Data were acquired on a 3\,T Siemens Trio scanner with a 32-channel head coil. Functional images were T2$^*$-weighted gradient-echo echo-planar images at 2.5\,mm isotropic resolution with a repetition time of 1.75\,s, an echo time of 30\,ms, a flip angle of 71$^\circ$, 54 slices, and a multi-band acceleration factor of 2. A T1-weighted anatomical image at 1\,mm isotropic resolution was acquired as the anatomical reference, and dual-echo field maps were acquired at the beginning of every session to correct the spatial distortion of the functional images.

\paragraph{Preprocessing.}
We use the release of the BOLD Moments Dataset under the MOSAIC framework, which aggregates vision fMRI datasets under a shared preprocessing pipeline~\citep{lahner2025mosaic}. fMRIPrep preprocesses each subject and registers the data to the fsLR32k cortical surface~\citep{esteban2019fmriprep}, and GLMsingle estimates one beta per presentation from the preprocessed time series~\citep{prince2022glmsingle}. The betas of each session are normalized by the mean and standard deviation of the training videos presented in that session. Averaging the normalized betas over the repetitions of a video leaves one response per video at each vertex.

\paragraph{Noise Ceiling.}
The noise ceiling is the upper bound on how well any model can predict the response of a vertex. In the MOSAIC pipeline the noise ceiling is calculated from the ten presentations of each test video. The variance across the ten presentations of the same video is the noise variance $\sigma^2_{\mathrm{noise}}$, and the remaining variance of the response is the signal variance $\sigma^2_{\mathrm{signal}}$. The response we model is the average over $n = 10$ presentations, which reduces the noise variance to $\sigma^2_{\mathrm{noise}} / n$. The noise ceiling is the fraction of the variance of the averaged response that is signal,
\begin{equation}
\mathrm{NC} = \frac{\sigma^2_{\mathrm{signal}}}{\sigma^2_{\mathrm{signal}} + \sigma^2_{\mathrm{noise}} / n}.
\label{eq:nc}
\end{equation}

\paragraph{Atlas.}
The HCP-MMP atlas parcellates each hemisphere into 180 areas grouped into 22 divisions by function and anatomy, and those 22 divisions into six major systems~\citep{glasser2016}. One of the six is the Visual System, which comprises the five divisions. Primary Visual is V1, 1{,}618 vertices. Early Visual is V2, V3, and V4, 2{,}565 vertices. The Dorsal Stream is V6, V3A, V7, IPS1, V3B, and V6A, 1{,}116 vertices. The Ventral Stream is V8, FFC, PIT, VMV1, VMV2, VMV3, and VVC, 1{,}250 vertices. The MT+ Complex is MST, LO1, LO2, MT, PH, V4t, FST, V3CD, and LO3, 1{,}345 vertices.

\subsection{Feature Construction and Encoding Model}
\label{app:pipeline}
\paragraph{Sparse Random Projection.}
Every representation is flattened over tokens and channels and reduced to $d = 6004$ dimensions with a sparse random projection $\Phi$~\citep{li2006srp},
\begin{equation}
X^{\mathrm{obs}} = \Phi\big(c^{\mathrm{obs}}\big),
\qquad
X^{\mathrm{fut}}_{s,b} = \Phi\big(H^{\mathrm{fut}}_{s,b}\big),
\qquad
X^{\mathrm{rec}}_{s,b} = \Phi\big(H^{\mathrm{rec}}_{s,b}\big),
\label{eq:srp}
\end{equation}
each lying in $\mathbb{R}^{N \times d}$. The number of projections follows the Johnson--Lindenstrauss lemma, and $\epsilon = 0.1$ for the $N = 1{,}102$ videos gives $d = 6004$.

\paragraph{Ridge Regression.}
We follow the ridge regression and parameter selection of~\citet{lin2024stacking}. The encoding model of Equation~\ref{eq:ridge} is solved in closed form,
\begin{equation}
\hat{B} = (X^\top X + \lambda I)^{-1} X^\top Y, \qquad \hat{Y} = X\hat{B}.
\end{equation}
The penalty $\lambda$ is chosen for each vertex from 16 values spaced by factors of ten between $10^{-6}$ and $10^{9}$. The outer loop divides the 1{,}000 training videos into five folds and predicts each fold from a model fit on the remaining four, which yields an out-of-fold prediction for every training video. The inner loop divides those four folds again into five and chooses $\lambda$ at each vertex by the inner cross-validation error. The model is then refit on all training videos with the chosen $\lambda$.

\paragraph{Cell Selection.}
Let $g = (s,b)$ denote a grid cell and let $X_g$ be its feature matrix. Let $R^{2,\mathrm{oof}}_{m,p}(X)$ denote the out-of-fold $R^2$ at vertex $p$ in subject $m$. For a region $r$ in subject $m$ we select the cell $g^\star_{r,m}$ that maximizes that subject's own out-of-fold score, averaged over the vertices of the region,
\begin{equation}
g^\star_{r,m} = \arg\max_{g}\ \frac{1}{|r|}\sum_{p \in r} R^{2,\mathrm{oof}}_{m,p}\big(X_{g}\big),
\label{eq:cellsel}
\end{equation}
and use it for every vertex of $r$ in that subject. Both the ridge penalty and $g^\star_{r,m}$ come from the training folds, so the test videos take no part in model selection. The encoding accuracy reported at a cell is the mean over subjects of each subject's test score at the cell selected on their own training set.

\subsection{Variance Partitioning}
\label{app:vp}

We use variance partitioning to separate what each feature space explains uniquely from what they explain in common~\citep{visconti2026, lin2024stacking}. For a pair $a$ and $b$ their predictions are combined at each vertex into a joint model, with convex weights fit to the training responses,
\begin{equation}
\hat{Y}^{\mathrm{stack}} = w_a\hat{Y}^{a} + w_b\hat{Y}^{b},
\qquad
(w_a, w_b) = \arg\min_{w_a + w_b = 1,\ w \ge 0}\ \big\lVert Y_{\mathrm{train}} - (w_a\hat{Y}^{a}_{\mathrm{oof}} + w_b\hat{Y}^{b}_{\mathrm{oof}})\big\rVert^2 ,
\label{eq:stack}
\end{equation}
which has a closed-form solution at each vertex. Let $\tilde{R}^2_a$, $\tilde{R}^2_b$, and $\tilde{R}^2_{\mathrm{stack}}$ denote the normalized test scores of the two feature spaces and of their combination. The unique and shared contributions are
\begin{equation}
U_a = \tilde{R}^2_{\mathrm{stack}} - \tilde{R}^2_b, \qquad
U_b = \tilde{R}^2_{\mathrm{stack}} - \tilde{R}^2_a, \qquad
U_{ab} = \tilde{R}^2_a + \tilde{R}^2_b - \tilde{R}^2_{\mathrm{stack}} .
\label{eq:vp}
\end{equation}

\subsection{Spatiotemporal Skip Guidance}
\label{app:stg}

Spatiotemporal skip guidance (STG) is a sampling-time guidance for video diffusion models~\citep{hyung2025stg}. In its attention variant (STG-Attention) it forms a perturbed weak model by skipping the self-attention of one layer $b$ and pushes the generation path away from that weak model,
\begin{equation}
\epsilon = \epsilon_{\mathrm{CFG}} + w\big(\epsilon_{\mathrm{cond}} - \epsilon_{\mathrm{perturb}}\big),
\qquad
\epsilon_{\mathrm{CFG}} = \epsilon_{\mathrm{uncond}} + s\big(\epsilon_{\mathrm{cond}} - \epsilon_{\mathrm{uncond}}\big).
\label{eq:stg}
\end{equation}
We set $w = 1.5$ and the classifier-free guidance scale $s = 5.0$. Here $\epsilon_{\mathrm{cond}}$ and $\epsilon_{\mathrm{uncond}}$ are the predictions of the intact model under the conditional and the unconditional prompt, and $\epsilon_{\mathrm{perturb}}$ is the prediction under the conditional prompt with the self-attention of layer $b$ skipped. In the skipped layer the attention scores are replaced by the identity.

We apply STG while the AR model generates the 5\,s video that follows the 3\,s observed video. The unguided arm uses classifier-free guidance alone, and each of the other eight arms adds the STG term for one layer $b \in \{0, 5, 11, 16, 22, 28, 32, 38\}$.

\subsection{Human Preference Evaluation}
\label{app:humaneval}

\paragraph{Stimuli.}
For each of the 102 test videos the AR model generates the 5\,s continuation under nine arms, eight with STG applied at one layer each and one baseline that uses classifier-free guidance alone. Each video is 8\,s long, the 3\,s original video followed by the 5\,s continuation, rendered at $272 \times 272$ and 33\,fps without sound.

\paragraph{Trial Design.}
In each trial an observer watches two videos side by side and chooses the one they prefer. The nine arms give $\binom{9}{2} = 36$ pairs of arms, and an observer sees every pair four times, each time with a different source video, so an observer completes 144 trials. The left and right positions are balanced within each pair. Five observers completed the test, 720 trials in total.

\paragraph{Bradley--Terry Model.}
To rank the arms from the pooled choices we fit a Bradley--Terry model~\citep{bradley1952}, in which arm $i$ is chosen over arm $j$ with probability
\begin{equation}
P(i \succ j) = \frac{\pi_i}{\pi_i + \pi_j},
\label{eq:bt}
\end{equation}
where $\pi_i$ is the worth of arm $i$. We report the log-worth $\theta_i = \log \pi_i$ with the unguided arm anchored at $\theta = 0$. A pseudo-count of $0.1$ wins and $0.1$ losses between every pair of arms keeps the estimate finite.

\subsection{Statistical Testing}
\label{sec:stats}

Paired comparisons across the ten subjects use an exact two-sided sign-flip test, whose smallest attainable $p$ is $0.002$. The correlation between the layer profiles of encoding accuracy and of preference is the Pearson $r$ over the eight guided layers, with a two-sided $p$ from an exact permutation test. Intervals on the log-worth are 95\% intervals from 2{,}000 bootstrap resamples of observers.

\end{document}